%% file: main.tex
\documentclass[sigconf]{acmart}
\usepackage[dvipsnames, svgnames]{xcolor}
\usepackage{makecell}

\usepackage{float}
\usepackage{algorithm}
\usepackage{algorithmic}
\usepackage{caption}
\usepackage{indentfirst}
\usepackage{microtype}
\AtBeginDocument{%
  }

\setcopyright{acmlicensed}
\copyrightyear{2026}
\acmYear{2026}
\acmDOI{XXXXXXX.XXXXXXX}

\acmConference[RecSys '26]{Twentieth ACM Conference on Recommender Systems}{September 28--October 02, 2026}{Minneapolis, MN, USA}
\acmISBN{978-1-4503-XXXX-X/26/09}


\copyrightyear{2026}
\acmYear{2026}
\setcopyright{cc}
\setcctype{by}
\acmConference[RecSys '26]{20th ACM Conference on Recommender Systems}{September 27-October 02, 2026}{Minneapolis, MN, USA}
\acmBooktitle{20th ACM Conference on Recommender Systems (RecSys '26), September 27-October 02, 2026, Minneapolis, MN, USA}
\acmDOI{10.1145/3773078.3831877}
\acmISBN{979-8-4007-2284-4/2026/09}
\begin{document}

\title{Shape Your Feed: An LLM-based Agentic System for Conversational Recommendation}

\author{Ziyun Xu}
\authornote{Corresponding authors.}
\email{vincentxu@meta.com}
\affiliation{%
  \institution{Meta Platforms}
  \city{Menlo Park}
  \state{California}
  \country{USA}
}

\author{Bosen Ding}
\email{bosending@meta.com}
\affiliation{%
  \institution{Meta Platforms}
  \city{Menlo Park}
  \state{California}
  \country{USA}
}

\author{Yue Zhang}
\email{zhangyue@meta.com}
\affiliation{%
  \institution{Meta Platforms}
  \city{Menlo Park}
  \state{California}
  \country{USA}
}

\author{Ji Qi}
\email{jiqi@meta.com}
\affiliation{%
  \institution{Meta Platforms}
  \city{Menlo Park}
  \state{California}
  \country{USA}
}

\author{Qingyuan Song}
\email{qsong@meta.com}
\affiliation{%
  \institution{Meta Platforms}
  \city{Menlo Park}
  \state{California}
  \country{USA}
}

\author{Jizhou Huang}
\email{jacksonhuang@meta.com}
\affiliation{%
  \institution{Meta Platforms}
  \city{Menlo Park}
  \state{California}
  \country{USA}
}

\author{Liwei Wang}
\email{liweiw@meta.com}
\affiliation{%
  \institution{Meta Platforms}
  \city{Menlo Park}
  \state{California}
  \country{USA}
}

\author{Jeffrey Santelli}
\email{jsantelli@meta.com}
\affiliation{%
  \institution{Meta Platforms}
  \city{Menlo Park}
  \state{California}
  \country{USA}
}

\author{Yue Weng}
\email{yweng@meta.com}
\affiliation{%
  \institution{Meta Platforms}
  \city{Menlo Park}
  \state{California}
  \country{USA}
}

\author{Qichao Que}
\email{qichao@meta.com}
\affiliation{%
  \institution{Meta Platforms}
  \city{Menlo Park}
  \state{California}
  \country{USA}
}

\author{Zhenheng Yang}
\email{zhenheny@meta.com}
\affiliation{%
  \institution{Meta Platforms}
  \city{Menlo Park}
  \state{California}
  \country{USA}
}

\author{Junfeng Pan}
\email{panjunfeng@meta.com}
\affiliation{%
  \institution{Meta Platforms}
  \city{Menlo Park}
  \state{California}
  \country{USA}
}

\author{Linhong Zhu}
\authornotemark[1]
\email{linhongzhu@meta.com}
\affiliation{%
  \institution{Meta Platforms}
  \city{Menlo Park}
  \state{California}
  \country{USA}
}

\renewcommand{\shortauthors}{Xu et al.}

\input{abstract}


\maketitle

\input{intro}
\input{related}
\input{problem}
\input{system}
\input{alignment}
\input{eval}

\input{con}

\input{acknowledgement}

\clearpage 
\bibliographystyle{ACM-Reference-Format}
\bibliography{SYF}

\end{document}

%% file: abstract.tex
\begin{abstract}
Industrial recommendation systems predominantly adopt a passive ranking paradigm that infers user preferences from implicit behavioral signals (e.g., clicks, dwell time) rather than explicit, natural language inputs. As a result, users experience a persistent discrepancy between their explicit interests and what passive behavioral algorithms deliver, limiting their ability to express nuanced preferences or steer their feed in real time. To address this growing gap between how recommendations are optimized and how users wish to articulate their interests, we present Shape Your Feed (SYF), an LLM-based agentic recommendation framework that enables real-time, multimodal co-curation of content. SYF employs a three-tier architecture: (i) a Perception Flow that captures fine-grained user intent from text prompts, voice commands, and UI interactions; (ii) a Serving Flow that performs real-time agentic re-ranking and pruning of candidate items, grounded in a persistent Semantic Profile encoding evolving user preferences; and (iii) a Self-Evolution Flow that aligns system behavior with human judgments via Direct Preference Optimization (DPO) and an LLM-as-a-Judge ensemble.
Offline evaluations show that SYF’s alignment scoring module achieves 98.85\% accuracy, substantially improving over strong few-shot baselines. Large-scale online A/B experiments on production traffic further demonstrate that SYF improves feed relevance and user sentiment, indicating a practical and scalable path toward interactive, user-steerable recommendation in industrial settings.
\end{abstract}

%% file: intro.tex
\section{Introduction}
Personalized content recommendation is central to maintaining user engagement on modern social platforms, where users consume a fast-moving stream of multi-format content (e.g., posts, short videos, stories). Today's feed ranking systems achieve strong relevance through implicit behavioral signals~\cite{hu2008collaborative, rendle2009bpr, he2017neural, covington2016deep, cheng2016wide, naumov2019deep}. However, platform users frequently experience a persistent discrepancy between their explicit interests and what passive behavioral algorithms deliver. Resolving this gap requires recommendation systems to empower users to express nuanced preferences, understand why content surfaces, and steer their experience in real time~\cite{govea2024transparency}. This demand motivates a new class of \emph{conversational} and \emph{interactive} recommendation systems~\cite{jannach2021survey}, where users directly communicate with the underlying models through natural language (text or voice) or simplified UI actions, and the system responds with transparent, grounded content reflecting explicit intent.

Recent progress in Large Language Models (LLMs) has made conversational recommendation systems practical: LLMs interpret intent, resolve ambiguity, and maintain dialogue context, enabling precise preference matching compared to traditional query- or button-driven interfaces~\cite{friedman2023leveraging,wang2023llmrec}. Yet, deploying LLMs in feed environments introduces unique challenges. First, recommendation must remain \emph{catalog-grounded}, ensuring that generated outputs strictly correspond to valid items the platform can actually serve. Due to linguistic ambiguity and generative complexities~\cite{tonmoy2024hallucination, dziri2021neural}, unconstrained generation can easily lead to hallucinated items or out-of-catalog suggestions; thus, systems must tightly bound how communicative intent is mapped to the production inventory. Second, conversational interactions are inherently multi-turn with mixed intents—users often combine immediate goals ("fewer political reels") with long-term interests ("more cooking"). Ensuring multi-turn consistency requires \emph{explicit state tracking} and integration with robust ranking pipelines~\cite{kostric2024preference}. Finally, the system must be \emph{interpretable and controllable} at scale: users must easily view and modify inferred states, and their inputs must responsively alter the feed layout under tight production latency budgets~\cite{chen2012critiquing}. These constraints motivate an agentic, catalog-grounded architecture, which are goals we target with our proposed system.

Prior work ensures reliable conversational recommendation using hybrid retrieval-generation, schema-constrained prompts, or RL alignment frameworks (e.g., LLM-ConvRec~\cite{feng2023large}, collaborative retrieval~\cite{zhu2025crag}, and Rank-GRPO/ConvRec-R1~\cite{zhu2025rankgrpo}). While these methods reduce hallucinations and improve state tracking, they focus primarily on \emph{item recommendation} within static catalogs (e.g., movies or products). They fail to address the demands of \emph{high-throughput feed environments}, where ranking pipelines must continuously and responsively react to diverse, fine-grained feedback across multiple modalities and UI surfaces.

In this work, we introduce \textbf{Shape Your Feed (SYF)}, an LLM-based conversational recommendation system enabling \textbf{direct, real-time user control}. SYF unifies \textbf{multi-modal feedback} (text, voice, UI controls) for explicit preferences ("more posts from close friends") and dislikes ("less clickbait"). We leverage LLMs as a \emph{structured controller} to handle: (i) intent detection and normalization, (ii) persistent memory aggregation of user states, and (iii) dynamic \emph{re-ranking} grounded in the platform's retrievable inventory. By maintaining an editable representation of the user preference state and continuously projecting it into ranking features for candidate selection, SYF delivers an interpretable, user-steerable feed while preserving catalog grounding and delivery quality.

Our contributions are threefold:
\begin{enumerate}
    \item \textbf{System and architecture:} We propose a scalable architecture for conversational feed recommendation that combines LLM-based intent understanding with \textbf{structured, persistent memory} and \textbf{production-compatible ranking flows}, enabling stable multi-turn behavior and catalog-grounded responses.
    \item \textbf{Multi-modal preference supervision:} We demonstrate how \textbf{multi-modal, user-driven feedback} (text, voice, and UI controls) can be unified into a consistent preference state that improves personalization, transparency, and user-\hspace{0pt}perceived agency.
    \item \textbf{Empirical validation and insights:} Through extensive offline experiments and online A/B testing, we show that SYF improves feed relevance and engagement, distilling practical insights regarding open challenges and promising research directions.
\end{enumerate}

Through these results, SYF advances recommendation systems from passive personalization toward interactive, user-algorithm co-curation, offering a principled path to more transparent and controllable feed ranking in large-scale social platforms.

%% file: related.tex
\section{Related Works}
Our work builds upon and extends research across implicit feedback recommendation, conversational systems, and large language models (LLMs) for personalization.

\textbf{Implicit Feedback \& Deep Recommenders:} Modern recommenders rely on implicit signals (clicks, dwell time, shares) to infer preferences~\cite{hu2008collaborative, rendle2009bpr}. Paradigms like matrix factorization~\cite{hu2008collaborative}, BPR~\cite{rendle2009bpr}, and neural hybrid architectures (e.g., Wide \& Deep~\cite{cheng2016wide}, neural collaborative filtering~\cite{he2017neural}) paved the way for industrial scale models like YouTube's DNN~\cite{covington2016deep} and DLRM~\cite{naumov2019deep}. Sequential attention models like SASRec~\cite{kang2018selfattentive} and BERT4Rec~\cite{sun2019bert4rec} capture temporal dynamics, but these architectures rely on passive signals and offer limited mechanisms for explicit user steering, a limitation SYF directly addresses.

\textbf{Conversational Recommender Systems (CRS):} CRS allow users to express interests via active dialogue~\cite{jannach2021survey, gao2021advances}. Early designs leveraged slot-filling~\cite{louvan2020neural} or attribute critiquing~\cite{chen2012critiquing}, while datasets like ReDial~\cite{li2018towards} spurred end-to-end models optimizing both dialogue and items. To improve alignment, knowledge graphs have been fused into conversational pipelines~\cite{chen2019kgsf, zhou2020improving, wong2021improving, ren2024explicit} to bridge text tracking and inventory retrieval. However, most CRS assume static catalogs (e.g., books, movies) and fail to address high-throughput feed dynamics requiring real-time multi-modal adjustments and rapid negative constraint enforcement.

\textbf{LLMs for Recommendation:} LLMs introduce semantic intent understanding, reasoning, and explanation capabilities to personalization stacks~\cite{wu2023survey, fan2023recommender}. In conversational settings, RecLLM~\cite{friedman2023leveraging} integrates profile tracking on YouTube, and LLM-ConvRec~\cite{feng2023large} enhances response quality. Grounding inventory to mitigate hallucinations remains a primary challenge~\cite{tonmoy2024hallucination}. Retrieval-augmented frameworks like CRAG~\cite{zhu2025crag} couple LLM generation with collaborative filters, while RA-Rec~\cite{kemper2024retrieval} tracks state across turns. Alignment methods like Rank-GRPO~\cite{zhu2025rankgrpo} optimize list-wise rankings, and fine-tuning off-the-shelf LLMs to mimic normative Bayesian updates~\cite{qiu2026bayesian} improves multi-round probabilistic reasoning. We build on these by executing agentic list-wise refinement over high-throughput content streams.

%% file: problem.tex
\section{Problem Scope}
Modern industrial feed recommenders rely on a Passive Ranking paradigm~\cite{hu2008collaborative}. Formally, given a user $u \in \mathcal{U}$ and a candidate item $v \in \mathcal{V}$, the system estimates an engagement probability $\hat{y}_{u,v}$ (e.g., click or dwell time) based on the user's historical implicit behaviors $\mathcal{H}_u$ and current context $\mathcal{E}$:
\begin{equation}
    \hat{y}_{u,v} = f_{\theta}(v, \mathcal{H}_u, \mathcal{E})
\end{equation}
where $f_{\theta}$ is the production system parameterized by $\theta$. The fundamental limitation of this formulation is that it treats preference solely as a latent variable inferred from $\mathcal{H}_u$. Lacking a direct variable for conscious, real-time articulation of interest, it decouples the optimization objective from immediate user agency.

This structural limitation manifests in three major challenges:
\begin{itemize}
    \item \textbf{Black-box Nature:} Recommender logic is opaque to users. Because $\hat{y}_{u,v}$ is computed in a high-dimensional latent space, users lack explicit channels to understand specific recommendations or provide corrective feedback.
    \item \textbf{Coarse-grained Control:} Control mechanisms are limited to binary or categorical actions (e.g., clicking "Show Less" or dismissing a post). These updates to $\mathcal{H}_u$ are too coarse to distinguish whether a user dislikes the topic, creator, or tone.
    \item \textbf{Lack of Multi-modal Input:} Production systems optimize for single-tap interactions, lacking the capability to process multi-modal inputs like free-form text or voice commands. Without expressive channels, users cannot communicate complex intentions (e.g., "Show me more tech news but avoid AI hype"), forcing them to remain passive recipients.
\end{itemize}

LLMs resolve these bottlenecks through advanced semantic understanding and multi-modal support. This enables a shift from passive predictions to an \textit{Agentic System for Conversational Recommendation}. Integrating an agent into the feed stack addresses three primary objectives:
\begin{itemize}
    \item \textbf{Relevance to User Intent:} Natural language constraints allow the system to align content directly with explicit, real-time user intentions.
    \item \textbf{Explainability \& Transparency:} Providing clear reasons for recommendations makes system perception visible and adjustable, enhancing trust and user retention.
    \item \textbf{Overall Recommendation Quality:} Interactive feedback loops yield fine-grained user control, improving long-term satisfaction beyond the limits of standard production stacks.
\end{itemize}

%% file: system.tex
\section{System Overview}
\label{sec:system_overview}

\begin{figure*}
    \centering
    \includegraphics[width=\textwidth]{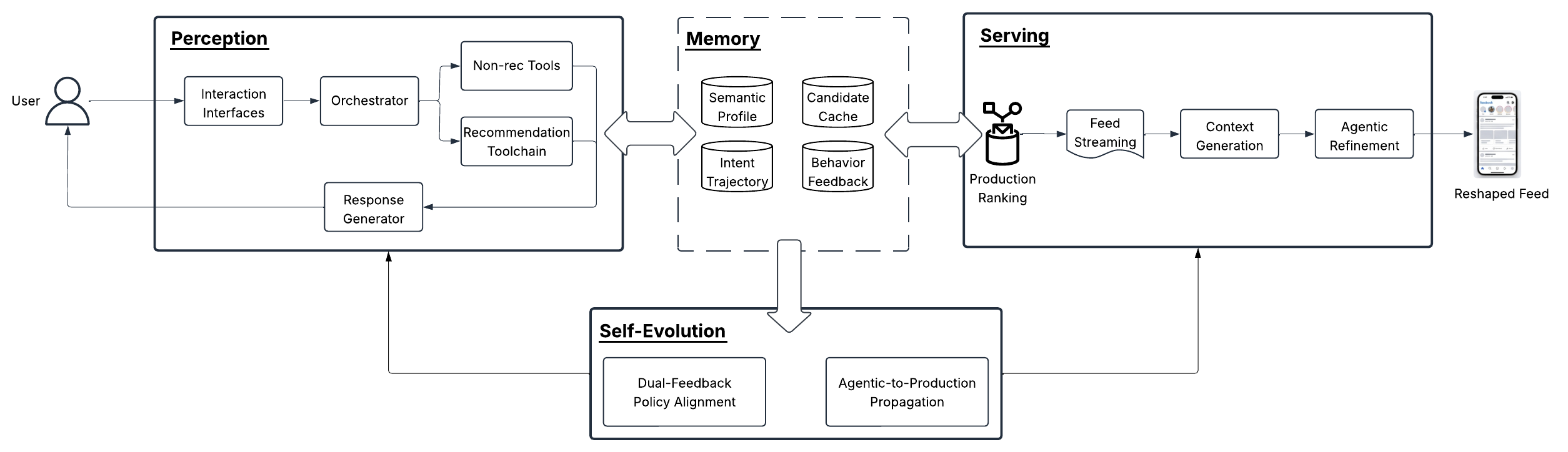} 
    \caption{\textbf{An Overview of Shape Your Feed System Architecture.}}
    \label{fig:system}
\end{figure*}

To address the limitations of passive ranking, we propose SYF, an agentic framework that redefines recommendation as a continuous, user-steerable personalization. The system architecture, illustrated in \textbf{Figure~\ref{fig:system}}, operates through three coupled flows—Perception, Serving, and Self-Evolution. 

While individual modules rely on standard NLP tuning (SFT and DPO) for late-stage alignment scoring, SYF's ``agentic'' branding reflects its closed-loop system architecture rather than a single model. Instead of acting as a static, single-turn classifier, it orchestrates a continuous state-tracking and execution loop. Specifically, SYF dynamically maintains a persistent Semantic Profile ($S_t$), monitors user intent trajectories ($\mathcal{T}_t$) across multi-turn interactions, and autonomously executes real-time serving-layer policies including candidate sourcing, pruning, and alignment. The fine-tuned models thus serve as the core decision engine within a broader, responsive, and self-evolving system.

\begin{enumerate}
    \item \textbf{Perception Flow (Section~\ref{sec:perception}):} Estimates user intent by resolving unstructured interactions into a semi-structured \textit{Semantic Profile} $S_t$ via the perception subsystem $\mathcal{A}_{\text{P}}$:
    \begin{equation}
        S_t = \mathcal{A}_{\text{P}}(\mathcal{I}_t, S_{t-1}, \mathcal{T}_{t-1})
    \end{equation}
    where $\mathcal{I}_t$ is the multi-modal interaction at time $t$, $S_{t-1}$ is the current profile, and $\mathcal{T}_{t-1}$ is the historical \textit{Intent Trajectory}. $\mathcal{A}_{\text{P}}$ coordinates an LLM-based modular workflow executing query generation, intent identification, and reasoning to synthesize inputs accurately.
    \item \textbf{Serving Flow (Section~\ref{sec:serving}):} Executes policy by projecting $S_t$ onto the content inventory to produce the final feed ranking $\mathcal{R}$ via serving agent $\mathcal{A}_{\text{S}}$:
    \begin{equation}
        \mathcal{R} = \mathcal{A}_{\text{S}}(\mathcal{V}_{final}, S_t; f_{\theta})
    \end{equation}
    where $\mathcal{V}_{final}$ is the final candidate pool blending production outputs with the perception-retrieved \textit{Candidate Cache}. Operating as a late-stage ranking layer over the production stack, $\mathcal{A}_{\text{S}}$ executes final feed distribution adjustments (augmentation, pruning, re-ranking) while utilizing base scores ($f_{\theta}$) to safeguard foundational engagement quality.
    \item \textbf{Self-Evolution Flow (Section~\ref{sec:evolution}):} Establishes bidirectional optimization. Through \textit{Dual-Feedback Policy Alignment}, it leverages online behaviors and offline LLM-as-a-Judge evaluations to iteratively update policies $\mathcal{A}_{\text{P}}$ and $\mathcal{A}_{\text{S}}$. Simultaneously, via \textit{Agentic-to-Production Propagation}, it distills high-order semantic insights to update the base model $f_{\theta}$ for progressive intent adaptation.
\end{enumerate}

\begin{algorithm}
\caption{Perception Agent Workflow ($\mathcal{A}_{\text{P}}$)}
\label{alg:perception_flow}
\small
\begin{algorithmic}[1]
\REQUIRE 
    $\mathcal{I}_t$: Multi-modal interaction at time $t$; \\
    $S_{t-1}$: Previous Semantic Profile; \\
    $\mathcal{T}_{t-1}$: Historical Intent Trajectory; \\
    $\mathcal{C}_{s, t-1}$: Existing Candidate Cache.
\ENSURE 
    $S_t$: Updated Semantic Profile; \\
    $\mathcal{M}_t$: Agent response message; \\
    $\mathcal{T}_t$: Updated Intent Trajectory; \\
    $\mathcal{C}_{s, t}$: Refreshed Candidate Cache.

\STATE $q_t \leftarrow \texttt{Preprocess}(\mathcal{I}_t)$
\STATE $\mathcal{K}_{tasks} \leftarrow \texttt{Orchestrator}(q_t)$
\STATE $\mathcal{O}_{tools} \leftarrow \emptyset$
\FORALL{$k \in \mathcal{K}_{tasks}$}
    \IF{$k \in \text{Non-Rec-Tools}$}
        \STATE $res_k \leftarrow \texttt{InvokeTool}(k, q_t)$
        \STATE $\mathcal{O}_{tools}.\text{add}(res_k)$
    \ELSE 
    \STATE \COMMENT{Recommendation Toolchain}
        \STATE $I_t \leftarrow \texttt{DetectIntent}(q_t)$
        \STATE $S_t \leftarrow \texttt{SynthesizeProfile}(q_t, S_{t-1}, I_t, \mathcal{T}_{t-1})$
        \STATE $\mathcal{T}_t \leftarrow \mathcal{T}_{t-1} \cup \{(q_t, I_t, S_t)\}$
        
        \STATE $\Delta S_t^+ \leftarrow \texttt{ExtractNewPositives}(S_t, S_{t-1})$
        
        \IF{$\Delta S_t^+ \neq \emptyset$}
            \STATE \COMMENT{Trigger asynchronously}
            \STATE \textbf{Spawn} $\mathcal{C}_{s, t} \leftarrow \texttt{SourceCandidates}(\Delta S_t^+, \mathcal{C}_{s, t-1})$
        \ENDIF
        
        \STATE $res_{rec} \leftarrow \texttt{GenerateConfirmation}(S_t)$
        \STATE $\mathcal{O}_{tools}.\text{add}(res_{rec})$
    \ENDIF
\ENDFOR

\STATE $\mathcal{M}_t \leftarrow \texttt{ResponseGenerator}(q_t, \mathcal{O}_{tools})$
\RETURN $\mathcal{M}_t, S_t, \mathcal{T}_t, \mathcal{C}_{s, t}$
\end{algorithmic}
\end{algorithm}

\subsection{Perception Flow} 

\label{sec:perception}
\begin{figure*}[t]
    \centering
    \includegraphics[width=\textwidth]{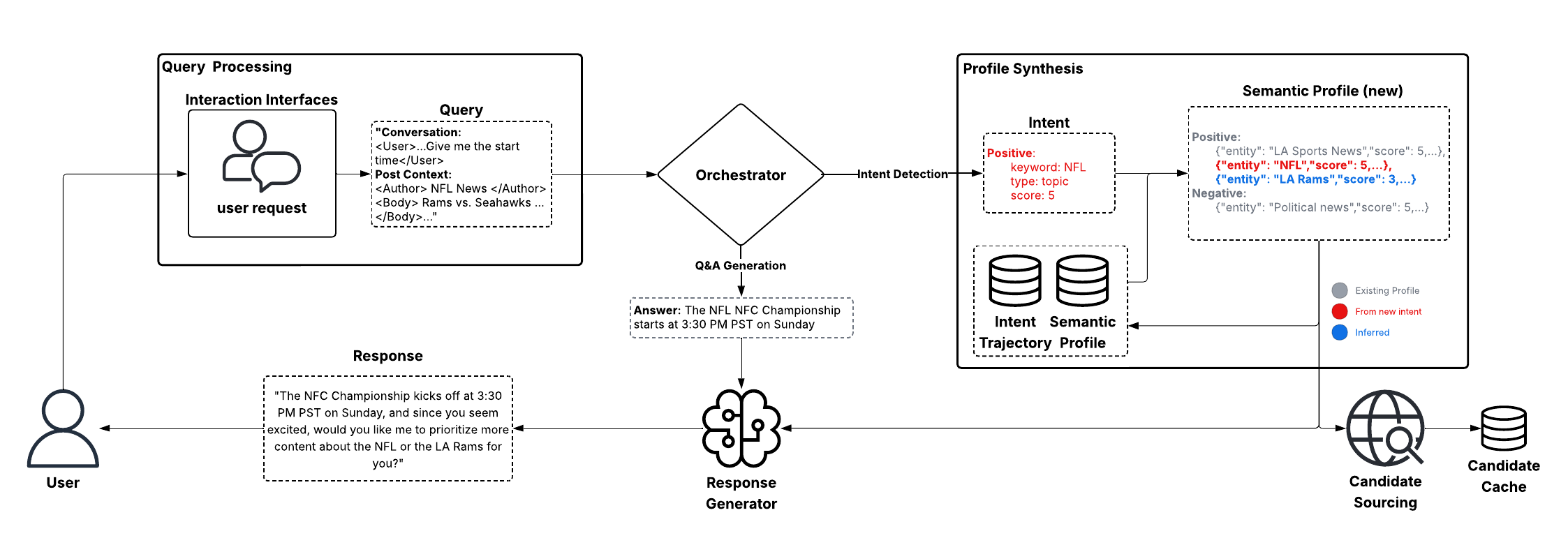} 
    \caption{\textbf{An Illustrative Example of the Perception Flow.}}
    \vspace{-0.4em}
    \label{fig:perception}
\end{figure*}

The Perception Flow $\mathcal{A}_{P}$ parses unstructured multi-modal interactions $I_{t}$ into a semi-structured Semantic Profile $S_{t}$. As shown in Figure~\ref{fig:perception} and Algorithm~\ref{alg:perception_flow}, this pipeline maps vague feedback to ranking constraints through four main stages: interaction interfaces, orchestrator, recommendation toolchain, and response generator.

\subsubsection{Interaction Interfaces} The system provides three multi-modal entry points within the host app to facilitate user control:
(1) \textbf{Context-Aware Feedback Pills}, where LLMs analyze a post's semantic context to generate granular reasoning options (sub-topics or content styles) for precise, one-tap feedback;
(2) \textbf{Content Preference Settings}, featuring a centralized homepage hub allowing users to proactively shape consumption by selecting topic bubbles or inputting free-form preferences; and
(3) \textbf{MetaAI Assistant Integration}, which acts as a conversational agent supporting text and voice dialogue to process user-assistant history and environmental context into a unified query $q_t$.

\subsubsection{Orchestrator} 
\begin{figure}
    \centering
    \includegraphics[width=\columnwidth]{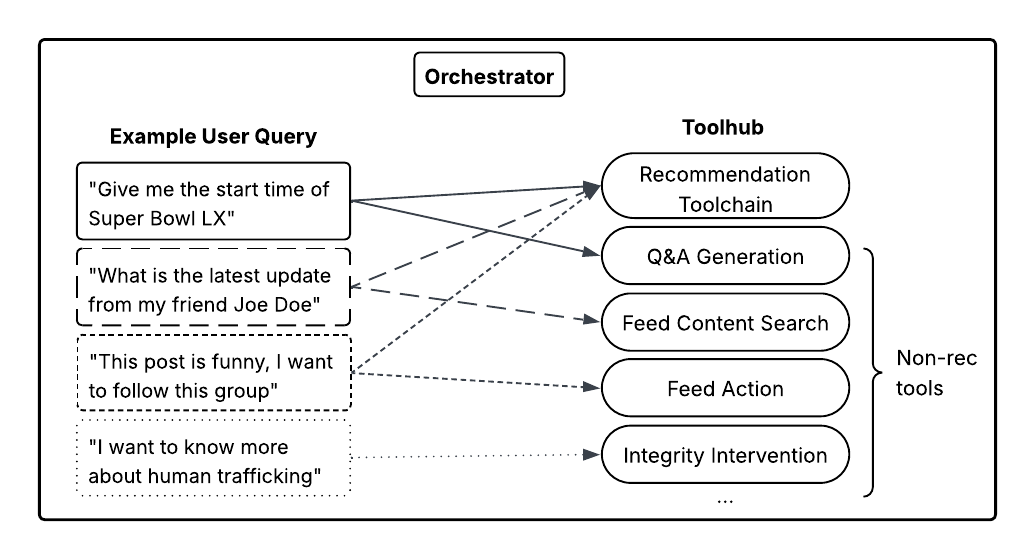} 
    \caption{\textbf{Orchestrator's Multitask Intent Parsing and Routing Logic.}}
    \label{fig:orchestrator}
\end{figure}

The Orchestrator parses unified queries $q_{t}$, extracts user intent, and routes tasks. Per Figure~\ref{fig:orchestrator}, it coordinates LLM reasoning and multi-tool invocation, triggering the \textit{Recommendation Toolchain} and non-recommendation tools (e.g., general Q\&A) simultaneously depending on intent complexity to ensure coherent profile updates. This work specifically focuses on the
Recommendation Toolchain and its impact on the feed recommen-
dation system.

\subsubsection{Recommendation Toolchain} 
A collection of LLM-based modular utilities invoked to track intent and update profiles:
\begin{enumerate} 
    \item \textbf{Intent Detection:} Deciphers $q_{t}$ into explicit intent $I_t$ by disambiguating underlying behavioral motivations.
    \item \textbf{Preference Synthesis:} Consolidates $I_t$, current profile $S_{t-1}$, and history $\mathcal{T}_{t-1}$ into an updated profile $S_{t}$ via four operations: \textit{De-duplication} (merging similar intents), \textit{Strengthening} (emphasizing recurring topics), \textit{Conflict Resolution} (removing contradictory records), and \textit{Semantic Expansion} (associating related interests).
    \item \textbf{Candidate Sourcing:} Asynchronously retrieves relevant inventory and executes lightweight ranking (Algorithm~\ref{alg:candidate_sourcing}) whenever new positive interests $\Delta S_t^+$ emerge. This asynchronous design decouples processing from the main serving path, minimizing latency costs.
\end{enumerate}

\subsubsection{Response Generator}
The Response Generator synthesizes query and tool outputs into a coherent message, dynamically adapting its modality to transparently confirm intent processing prior to the serving phase.

\begin{algorithm}
\caption{Async Candidate Sourcing ($\text{SourceCandidates}$)}
\label{alg:candidate_sourcing}
\small
\begin{algorithmic}[1]
\REQUIRE 
    $\Delta S_t^+$: New positive semantic interests; \\
    $\mathcal{C}_{s, t-1}$: Existing candidate cache. \\
    \textbf{System Resources \& Parameters:} \\
    $\mathcal{V}$: Content inventory; \\
    $\mathcal{H}_u$: User historical behaviors; \\
    $\mathcal{E}$: Current context; \\
    $f'_{\theta}$: Lightweight ranking model; \\
    $K$: Retrieval threshold per interest.
    
\ENSURE $\mathcal{C}_{s, t}$: Refreshed Candidate Cache.

\STATE $\mathcal{V}_{new} \leftarrow \emptyset$ \COMMENT{Initialize container for new candidates}

\FORALL{$i^+ \in \Delta S_t^+$}
    \STATE $\mathcal{C}_{raw} \leftarrow \texttt{InventorySearch}(i^+, \mathcal{V})$
    \STATE $\mathcal{C}_{scored} \leftarrow \emptyset$
    \FORALL{$v \in \mathcal{C}_{raw}$}
        \STATE $\hat{y}_{u,v} \leftarrow f'_{\theta}(v, \mathcal{H}_u, \mathcal{E})$ \COMMENT{Lightweight inference}
        \STATE $\mathcal{C}_{scored}.\text{add}((v, \hat{y}_{u,v}))$
    \ENDFOR
    \STATE $C_{i^+} \leftarrow \texttt{SelectTopK}(\mathcal{C}_{scored}, K, \text{by } \hat{y}_{u,v})$
    
    \STATE $\mathcal{V}_{new}.\text{add}(\{i^+: C_{i^+}\})$ \COMMENT{Map interest to top items}
\ENDFOR

\STATE $\mathcal{C}_{s, t} \leftarrow \mathcal{C}_{s, t-1} \cup \mathcal{V}_{new}$ \COMMENT{Merge with existing cache}

\RETURN $\mathcal{C}_{s, t}$
\end{algorithmic}
\end{algorithm}

\subsection{Serving Flow}
\label{sec:serving}
\begin{figure*}
    \centering
    \includegraphics[width=\textwidth]{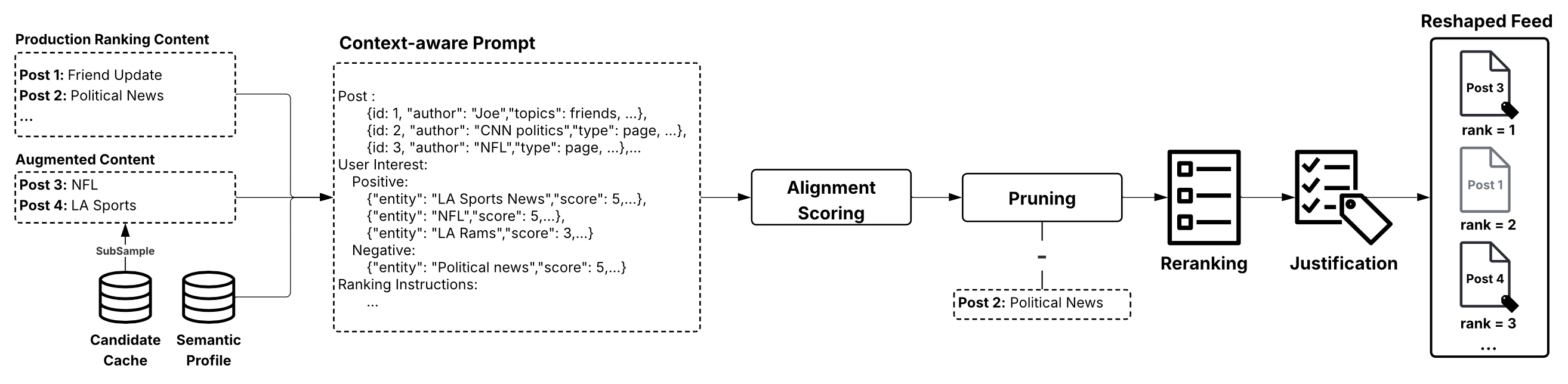} 
    \caption{\textbf{An Illustrative Example of the Serving Flow.}}
    \label{fig:serving}
    \vspace{-0.3em}
\end{figure*}

The Serving Flow (Algorithm~\ref{alg:serving_flow}, Figure~\ref{fig:serving}) manages real-time feed modifications, embedding agentic control into traditional ranking to maximize alignment with explicit user intent.

\subsubsection{Context Generation} This module is responsible for translating numerical ranking signals into a semantic format interpretable by the LLM. In production environments, item features are typically represented as hashed IDs or dense vectors optimized for efficiency. This module decodes these raw attributes into natural language descriptions (e.g., topic strings, creator metadata). It then synthesizes these textual features with the user's Semantic Profile $S_t$ to construct a comprehensive context-aware prompt $\mathcal{P}_{align}$. This prompt serves as the input for the subsequent \textit{Agentic Refinement} stage, effectively bridging the gap between the system's structural ranking signals and the user's explicit semantic intentions.

\subsubsection{Agentic Refinement} This module drives late-stage LLM intervention via $\mathcal{P}_{align}$ to execute sequential refinement operations (Algorithm~\ref{alg:serving_flow}):
\begin{itemize}
    \item \textbf{Augmentation:} Subsamples high-relevance items from the Candidate Cache ($\mathcal{C}_{s,t}$) to form $\mathcal{C}_{aug}$, merging them with production candidates ($\mathcal{C}_{prod}$) into an augmented pool $\mathcal{C}_{pool} = \mathcal{C}_{prod} \cup \mathcal{C}_{aug}$.
    \item \textbf{Alignment Scoring:} Evaluates list-wise inference scores $\hat{y}_{v|S_t}$ across $\mathcal{C}_{pool}$ against $\mathcal{P}_{align}$. This scoring is strictly bidirectional: candidates matching positive interests receive positive values, while those aligning with negative constraints (e.g., disliked topics) are assigned negative scores.
    \item \textbf{Pruning:} Applies a hard filter where any candidate $v$ scoring below a threshold $\tau$ is strictly purged from $\mathcal{C}_{pool}$, enforcing negative feedback instantly.
    \item \textbf{Re-ranking:} Blends semantic alignment scores with production engagement predictions ($\hat{y}_{u,v}$) to determine a final sorting metric $s_{final}$:
    \begin{equation}
        s_{final} = \alpha \cdot \hat{y}_{v|S_t} + (1-\alpha) \cdot \hat{y}_{u,v}
    \end{equation}
    where $\alpha$ controls fusion weight, preserving baseline engagement quality while tracking user intent. This explicitly derived score effectively surfaces relevant content for cold-start interests or low-signal users with weak historical priors.
    \item \textbf{Justification:} Generates concise, natural language explanations for top items (e.g., "Because you asked for more [Topic X]") to reinforce feed transparency.
\end{itemize}

\subsection{Self-Evolution Flow}
\label{sec:evolution}
The Self-Evolution Flow drives closed-loop optimization to synchronize knowledge across models and refine agent policies, utilizing interaction records in the \textit{Memory Persistence Layer} for dual-path tuning: optimizing the agent's reasoning while back-propagating signals to enhance the production ranking stack.

\subsubsection{Agentic-to-Production Propagation}
To augment the production ranker $f_{\theta}$, high-order latent intents from the LLM agent are distilled into structured priors across three vectors:
(1) \textbf{Representation Enhancement}, where $S_t$, $\mathcal{T}_t$, and $\mathcal{H}_{u, \mathcal{A}}$ serve as features to capture user-agent-feed interaction dynamics;
(2) \textbf{Retrieval Augmentation}, which integrates $S_t$ into initial retrieval to prioritize explicit semantic intents early in the funnel; and
(3) \textbf{Late-stage Personalization}, which leverages $S_t$ for granular ranking adjustments.
Further domain-specific optimization details are omitted as they lie outside the core SYF framework.

\subsubsection{Dual-Feedback Policy Alignment}
Runs an automated train--evaluate--deploy loop combining sparse, high-fidelity live Behavioral Feedback $\mathcal{H}_{u, \mathcal{A}}$ as ground-truth anchors with an offline LLM-as-a-Judge ensemble serving as a coverage amplifier to ensure high-throughput alignment scaling. Section~\ref{sec-alignment} outlines implementation details.

\subsection{Memory Persistence Layer}
The Memory Persistence Layer serves as the centralized data backbone, maintaining both persistent and transient states while synchronizing with the Perception, Serving, and Self-Evolution flows:
(1) \textbf{Semantic Profile ($S_t$)}, a semi-structured repository verbalizing fine-grained interests, disinterests, and intent strengths textually to support serving-layer reasoning;
(2) \textbf{Intent Trajectory ($\mathcal{T}_t$)}, which records temporal intent transitions to preserve longitudinal interaction context across multiple user sessions;
(3) \textbf{Candidate Cache ($C_{s,t}$)}, which asynchronously accumulates high-potential inventory items from positive interest updates ($\Delta S_t^+$), decoupling intensive retrieval loops from the serving track; and
(4) \textbf{Behavioral Feedback ($\mathcal{H}_{u,\mathcal{A}}$)}, which logs multi-modal user actions and dialogue choices across perception and serving layers for self-evolution reward optimization.

\begin{algorithm}
\caption{Serving Agent Workflow ($\mathcal{A}_S$)}
\small
\label{alg:serving_flow}
\begin{algorithmic}[1]
\REQUIRE 
    $S_t$: User Semantic Profile; \\
    $\mathcal{C}_{s, t}$: Candidate Cache from perception flow; \\
    $\mathcal{C}_{prod}$: Production Candidates with scores $\{(v, \hat{y}_{u,v})\}$; \\
    $\tau, \alpha$: Pruning threshold and fusion weight.
\ENSURE 
    $\mathcal{R}$: Final Ranked Feed.

\STATE $\mathcal{C}_{aug} \leftarrow \texttt{Subsample}(\mathcal{C}_{s, t})$
\STATE $\mathcal{C}_{pool} \leftarrow \mathcal{C}_{prod} \cup \mathcal{C}_{aug}$
\STATE $\mathcal{P}_{align} \leftarrow \texttt{GenerateContext}(S_t, \mathcal{C}_{pool})$
\STATE $\mathcal{S}_{align} \leftarrow \texttt{LLMScoring}(\mathcal{P}_{align})$
\STATE $\mathcal{C}_{temp} \leftarrow \emptyset$

\FORALL{$(v, \hat{y}_{u,v}) \in \mathcal{C}_{pool}$}
    \STATE $\hat{y}_{v|S_t} \leftarrow \mathcal{S}_{align}[v]$
    \IF{$\hat{y}_{v|S_t} < \tau$}
        \STATE \COMMENT{Prune items matching negative intent}
        \STATE \textbf{continue} 
    \ENDIF
    \STATE $s_{final} \leftarrow \alpha \cdot \hat{y}_{v|S_t} + (1-\alpha) \cdot \hat{y}_{u,v}$
    \STATE $\mathcal{C}_{temp}.\text{add}((v, s_{final}))$
\ENDFOR

\STATE $\mathcal{R} \leftarrow \texttt{Sort}(\mathcal{C}_{temp}, \text{by } s_{final})$
\RETURN $\mathcal{R}$
\end{algorithmic}
\end{algorithm}

%% file: alignment.tex
\section{Dual-Feedback Policy Alignment}\label{sec-alignment}
SYF contains multiple LLM-driven components optimized independently based on their respective objectives and supervision availability. We focus on the alignment scoring module optimization as a representative case study because it interfaces directly with explicit user feedback (e.g., clicking “Show Less” or dismissing a post), offering a robust signal for iterative refinement. For modules lacking explicit user signals, we apply prompt engineering and offline validation via an ensemble of LLM judges.

The alignment scoring component must satisfy two stringent production constraints:
\begin{itemize}
    \item \textbf{Latency Budgets:} Positioned on the critical execution path of high-throughput feeds, scoring must meet strict tail-latency bounds ($p_{99} < \delta$ ms), severely limiting the per-candidate computational budget.
    \item \textbf{Sparse Supervision:} Explicit positive or negative user actions are highly informative but naturally sparse, making purely feedback-based post-training unstable.
\end{itemize}

To satisfy the latency constraint, we adopt a \emph{list-wise approach}: given a user's Semantic Profile $S_t$ and candidates $\{v_i\}_{i=1}^K$, the model outputs alignment scores $\{\hat{y}_{v_i|S_t}\}_{i=1}^K$ in a single inference step, avoiding expensive point-wise scoring. To counter sparse feedback, we leverage an ensemble of open-source LLM judges (e.g., Llama 4 Maverick, Qwen3-VL-235B-A22B) to annotate $\langle S_t, v_i \rangle$ pairs where explicit signals are missing, applying majority voting to ensure fidelity. These LLM judges achieved a 96\% agreement rate with human judgments on a multi-reviewed benchmark. 

\begin{figure}[t]
  \centering
  \includegraphics[width=\columnwidth]{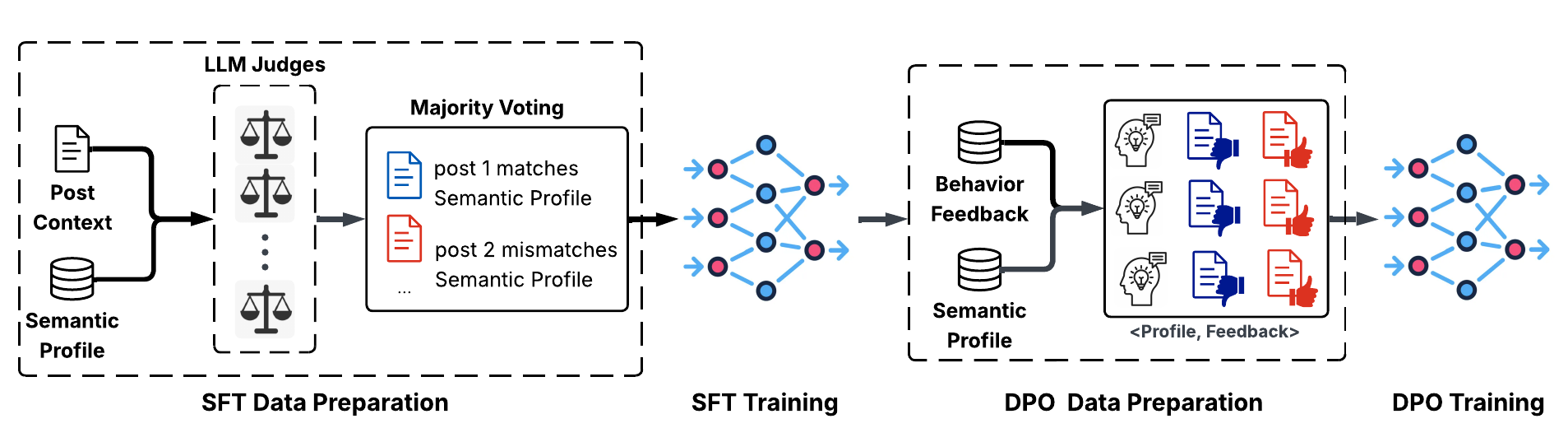}
  \caption{\textbf{Overview of the Dual-Feedback Policy Alignment.}}
  \label{fig:sft_dpc}
\end{figure}

We optimize alignment scoring in two stages. First, we establish a base policy via Supervised Fine-Tuning (SFT)~\cite{ouyang2022training} (Section~\ref{subsec-SFT}) using labels generated by the LLM judges. Second, we apply Direct Preference Optimization (DPO)~\cite{rafailov2023direct} (Section~\ref{subsec-DPO}) to refine the model using real user actions extracted from Behavioral Feedback $\mathcal{H}_{u, \mathcal{A}}$. Figure~\ref{fig:sft_dpc} provides a comprehensive overview of this dual-feedback policy alignment framework. 

\subsection{SFT Stage}\label{subsec-SFT}
The SFT stage establishes a robust foundational policy capturing domain-specific nuances and output constraints, serving as a warm-start for subsequent DPO optimization.

\subsubsection{SFT Data Preparation}
The SFT dataset is constructed by sampling historical feed-serving contexts from user engagement histories. Each training instance contains a user's current Semantic Profile and a set of feed candidates. Because explicit user feedback is limited, labels are generated via the LLM-as-a-Judge ensemble. A concise rationale for the winning label is generated by prompting Llama 4 Maverick to serve as the reasoning path. The evaluation set is built using the same pipeline to guarantee a consistent offline benchmark distribution. Crucially, explicit user feedback is omitted during SFT and reserved solely as direct supervision for DPO.

\subsubsection{Training Configuration}
We fine-tune a Llama3-8B model on a single machine with 8$\times$80GB GPUs. Hyperparameters are set to a batch size of 1, model parallelism of 1, and a learning rate of $2e^{-5}$. Training requires approximately 2.5 GPU hours per epoch.

\subsection{DPO Stage}\label{subsec-DPO}
DPO continues optimizing the alignment scoring using chosen-rejected preference pairs derived from both explicit user feedback contexts and judge-annotated contexts where direct signals are absent. We run DPO optimization in a recurring fashion to enable continuous improvement without a separate reward model or online RL rollouts, iteratively hardening the preference boundaries.

\subsubsection{DPO Data Preparation}
The DPO dataset maps preference pairs over organic contents. To augment sparse explicit negative labels, the LLM judge ensemble infers chosen–rejected comparisons when direct user feedback is unavailable. Concise explanations supporting the user feedback and winning LLM judge labels are generated by prompting
Llama 4 Maverick and serve as the reasoning path. We maintain the exact SFT evaluation distribution to isolate gains attributable to preference optimization. 

\subsubsection{Training Configuration}
DPO training utilizes 2 machines each equipped with 8$\times$A100 80GB GPUs, a batch size of 4, model parallelism of 1, and a learning rate of $2e^{-6}$ for 1 full epoch.

%% file: eval.tex
\section{Evaluation}
\subsection{Offline Evaluation}
We evaluate the performance of SYF's core LLM elements, focusing strictly on the execution metrics of the alignment scoring module. 
\subsubsection{Evaluation Metrics.} We leverage precision, recall, accuracy, and f1-score to assess the effectiveness of this module in the offline setting.
\subsubsection{Comparison to Few-shot baseline.}
To evaluate our method, we benchmark against a few-shot Llama3-8B baseline using an evaluation dataset annotated by our LLM judge ensemble. As shown in Table~\ref{tab:my_label}, the SFT candidate significantly outperforms the few-shot baseline across all dimensions, particularly in precision. The SFT + DPO configuration provides an additional 2.5\% f1-score improvement over the baseline SFT model.
\begin{table}
    \centering
    \caption{Effectiveness against Few-Shot Baseline.}
    \label{tab:my_label}
    \resizebox{\columnwidth}{!}{
    \begin{tabular}{lccccc} %
        \toprule
        \textbf{Method} & \textbf{Accuracy} & \textbf{Precision} & \textbf{Recall} & \textbf{f1-score}& \textbf{latency (ms)}\\
        \midrule
        Few-shot baseline & 83.84\% & 47.06\% & 60.22\% & 53.00\% & 650\\
        SFT               & 98.21\% & 78.20\% & 74.86\% & 76.97\% & 325\\
        SFT+DPO& 98.85\%& 79.16\%& 79.41\%& 79.51\%& 323\\
        \bottomrule
    \end{tabular}
    }
\end{table}
\subsubsection{Comparison to Conventional Ranking Model Baselines}
To evaluate whether conventional production rankers can capture explicit, real-time user intentions, we benchmark SYF against the platform's standard ranking signals: the predicted probability of a user clicking `Show More' ($P(\text{CSM})$) and `Show Less' ($P(\text{CSL})$). These standard signals are optimized via passive, implicit engagement histories. 

As shown in Table 2, when evaluating content segments that explicitly match versus mismatch the user's expressed Semantic Profile ($S_t$), the conventional ranking probabilities show negligible delta. For example, the 90th percentile of $P(\text{CSL})$ is virtually identical between matched and mismatched content ($0.208$ vs. $0.199$, a delta of only $0.009$). Similarly, $P(\text{CSM})$ fails to provide a discriminative margin ($0.289$ vs. $0.212$). 

Conversely, we evaluate the capability of our LLM alignment scoring module to act as a binary classifier of semantic alignment (classifying whether a feed item strictly matches or mismatches the user's profile $S_t$). The supervised fine-tuning (SFT) and SFT+DPO configurations show highly discriminative accuracy, especially SFT + DPO achieving $74.9\%$ and
$99.4\%$ accuracy, respectively.
\begin{table}
    \centering
    \caption{Effectiveness against $P(CSL)$ and $P(CSM)$.}
    \label{tab:re-ranking_effectiveness}
    \resizebox{\columnwidth}{!}{%
    \begin{tabular}{lcccc}
        \toprule
        \textbf{Condition} & \textbf{$P_{90}(\text{CSL})$}& \textbf{$P_{90}(\text{CSM})$}&\textbf{Accuracy SFT}& \textbf{Accuracy SFT + DPO}\\
        \midrule
        Feeds match user pref& 0.208& 0.289&0.749&  0.794\\
        Feeds mismatch user pref& 0.199& 0.212&0.992&  0.994\\
        \bottomrule
    \end{tabular}%
    }
    \small
    \raggedright
    \textit{Note: $P_{90}(\text{CSL})$ denotes 90 percentile of $P(CSL)$. $P_{90}(\text{CSM})$ denotes 90 percentile of $P(CSM)$.}
\end{table}

\subsubsection{Independent Human Evaluation}

To ensure the automated evaluation pipeline does not introduce systemic bias, we evaluated the model on an independent golden test set comprising 1029 human annotated samples. These samples were strictly isolated from the LLM judge pipeline. The system demonstrated highly comparable performance to the automated benchmarks, achieving an accuracy of 95.8\%, precision of 75.6\%, recall of 74.7\%, and an F1 score of 75.1\%. This tight alignment with human expert judgment confirms that our supervision framework effectively captures true semantic relevance without overfitting to synthetic artifacts.

\subsubsection{Base Model Selection}

We selected Llama 3 8B to optimally balance reasoning capability and production constraints. As shown in Table \ref{tab:model_selection}, the 1B model exhibits severely degraded performance and lacks the capacity for complex semantic alignment. Conversely, while the 70B variant improves precision, its massive memory footprint and inference latency prohibit deployment on the critical serving path. The 8B model provides sufficient accuracy while staying within the latency limits of our infrastructure.

\begin{table}
\centering
\caption{Zero Shot Performance Comparison Across Model Scales on the Golden Set}
\label{tab:model_selection}
\resizebox{\columnwidth}{!}{%
\begin{tabular}{lcccc}
\toprule
\textbf{Model} & \textbf{Accuracy} & \textbf{Precision} & \textbf{Recall} & \textbf{F1} \\
\midrule
Llama 3 8B (SFT+DPO) & 95.8\% & 75.6\% & 74.7\% & 75.1\% \\
Llama 3 70B & 94.6\% & 78.2\% & 49.4\% & 60.6\% \\
Llama 3 8B & 87.0\% & 32.3\% & 49.4\% & 39.1\% \\
Llama 3.2 1B & 56.9\% & 7.8\% & 37.9\% & 12.9\% \\
\bottomrule
\end{tabular}%
}
\end{table}

\subsection{Online Evaluation: System-Wide Performance}
\label{subsec:online_eval}
\subsubsection{Experiment Setup}
\label{subsubsec:online_setup}
We conducted a large-scale online A/B test on a randomized subset of live production traffic (US/Canada users, aged 18+) spanning several months. This duration ensured user adoption reached a steady state, outlasting the semantic memory retention window and neutralizing novelty anomalies. To isolate pure algorithmic updates from interface variations, interactions were restricted exclusively to Context-Aware Feedback Pills. The control group utilized the production ranking stack with static feedback options (e.g., "spam"), whereas the treatment group evaluated the proposed active SYF pipeline.
\subsubsection{Findings and Discussion}
\label{subsubsec:online_findings}
Online deployments show that SYF significantly enhances both user control interaction efficiency and overall feed relevance relative to production models. We detail the key findings below:
\paragraph{\textbf{Enhanced User Control Efficiency}} As shown in Table~\ref{tab:pill_distribution}, users demonstrated a definitive preference for Context-Aware Feedback Pills, which captured 76.02\% of total interactions. Given equal exposure, identical visual styling, and unbiased placement alongside static choices, this dominant usage shift underscores the superior semantic alignment of our LLM-generated options. By accurately parsing real-time intent, the agentic module creates an intuitive and responsive control mechanism.
\paragraph{\textbf{Reduction in Disliked Content}} We observed a marked decrease in explicit negative feedback, validating the system's effectiveness in filtering content that triggers user aversion. Specifically, per Table~\ref{tab:online_performance}:
\begin{itemize}
    \item The Overall Post Dismiss Rate dropped by 1.70\%.
    \item The Overall Post Dislike Rate fell by 2.74\%.
\end{itemize}
These metrics are reliable high-confidence proxies for tracking negative user experience in Feed Recommendation. Their simultaneous drop validates the efficacy of our Serving Flow, particularly the Pruning and Re-ranking, which block candidates matching negative interests stored in the Semantic Profile to enhance overall feed value.
\paragraph{\textbf{Improved Interest Exploration}} We achieved a statistically significant 0.16\% uplift in new interest consumption, a metric tracking the successful acquisition and retention of users within novel domains. This validates SYF's capacity for exploration, demonstrating that the architecture successfully uncovers latent intentions and guides the feed to support stable consumption patterns that historically constrained ranking models overlook.
\begin{table}
\centering
\small
\caption{Distribution of User Interactions with Pill Types.}
\label{tab:pill_distribution}
\begin{tabular}{lc}
\toprule
\textbf{Pill Type} & \textbf{Selection Share} \\
\midrule
Static Pills & 23.98\% \\
Context-Aware Pills & \textbf{76.02\%} \\
\bottomrule
\end{tabular}
\end{table}
\begin{table}[t] 
\centering
\small 
\caption{Feed Quality Improvement of the SYF System Relative Lift to the Production Baseline.}
\label{tab:online_performance}
\setlength{\tabcolsep}{12pt} 
\begin{tabular}{l r}
\toprule
\textbf{Metric} & \textbf{Full SYF System} \\
\midrule
Post Dismiss & \textbf{-1.70\%} \\
Post Dislike & \textbf{-2.74\%} \\
Interest Consumption & \textbf{+0.16\%} \\
\bottomrule
\end{tabular}
\end{table}

\subsubsection{Ablation Study}
\begin{table}
\centering
\caption{Ablation Study Results Relative to Production.}
\label{tab:ablation}
\begin{tabular}{lcc}
\toprule
\textbf{Metric} & \textbf{Pills Only} & \textbf{Full System} \\
\midrule
UI Click & +3.30\%* & +2.90\%* \\
Post Dismiss & +0.68\% & -1.70\%* \\
Post Dislike & -0.13\%  & -2.74\%* \\
Interest Consumption & -0.029\%  & +0.16\%* \\
\bottomrule
\multicolumn{3}{l}{\small * indicates a statistically significant difference ($p < 0.05$).}\end{tabular}
\end{table}

To isolate the contributions of the interface design and the underlying agentic execution, we conducted an online ablation study comparing the full system against a control group that deployed only the context-aware feedback pills interface without other components. As detailed in Table \ref{tab:ablation}, pills drive user interaction (+3.30\%* UI clicks) but yield no significant feed quality improvement alone. Dismiss and dislike capture all negative feedback including unintentional behavior. Without algorithmic reranking, these rates are not reduced. Similarly, without candidate augmentation, interest consumption does not increase. Conclusively, the pills improve signal collection while the serving flow improves signal execution.

\subsection{System Efficiency}
\label{sec:system_efficiency}

We optimize computational efficiency independently across the three primary components to balance execution speed with modeling capacity:

\begin{itemize}
    \item \textbf{Perception Flow:} The Perception Flow accommodates higher latency tolerance and lower query volume. Within this flow, asynchronous candidate sourcing decouples retrieval operations from the real-time execution path to effectively hide processing delays.
    \item \textbf{Serving Flow:} The Serving Flow handles high query volumes under strict latency constraints. It mitigates overhead by employing a smaller fine-tuned model that utilizes list-wise inference to score all candidates in a single computational step.
    \item \textbf{Self-Evolution Flow:} The Self-Evolution Flow operates entirely offline. Resource consumption in this stage is restricted solely to teacher model inference queries and standard training computation.
\end{itemize}

Production latency is influenced by many factors: multiple ranking pipelines trigger in parallel with sync/async execution. Through extensive optimization, SYF's overhead is largely hidden — adding +0.043\% critical path latency and +0.412\% direct tier latency. The primary contributor is alignment scoring inference (323ms, Table 1); other components integrate to the existing ranking pipeline with minimal impact. 

%% file: con.tex
\section{Conclusion}

In this work, we presented \textbf{Shape Your Feed (SYF)}, an agentic framework that bridges the gap between passive behavioral ranking and active user steering. Addressing the challenges of interpretability and real-time control introduced at the outset, SYF leverages a modular architecture---comprising Perception, Serving, and Self-Evolution flows---to ground expressive LLM reasoning within a high-throughput industrial environment. Extensive offline evaluations validate that our Dual-Feedback Policy Alignment mechanism effectively utilizes sparse signals to significantly boost the LLM's re-ranking performance over few-shot baseline. Furthermore, large-scale online A/B experiments on production traffic confirm that SYF improves engagement and interest exploration while significantly reducing explicit negative feedback. 

Despite these strong results, a key limitation of the current framework is its reliance on explicit user-provided inputs to trigger the active personalization loop. Because the ratio of users who proactively offer natural language or UI feedback is naturally limited in high-throughput environments, passive consumers who rarely interact with steering controls will primarily default to baseline ranking. Future work will focus on mitigating this input sparsity by introducing proactive preference elicitation during critical user journey shifts, and exploring transfer learning techniques to propagate inferred semantic profiles to low-signal users.

Ultimately, SYF validates that shifting from predicting preference to co-curating with the user is both technically feasible and beneficial, offering a scalable paradigm for the next generation of conversational recommendation systems.

%% file: acknowledgement.tex
\section{Acknowledgments}
We express our heartfelt gratitude to Fei Sha for his exceptional guidance and critical review, which significantly shaped the direction of this work. We extend our gratitude to Chiio Tut, Parmeet Singh Bhatia, Hao Yan, Wanqiang Chen,  for their essential contributions to the software development, which were instrumental in bringing the final system to realization. We also thank Momo Jiao for exceptional design support, alongside Facundo Severi and Dwij Garg for vital product support.